\documentclass[runningheads]{llncs}
\PassOptionsToPackage{table}{xcolor}

\usepackage{eccvabbrv}

\usepackage{graphicx}
\usepackage{booktabs}

\usepackage[accsupp]{axessibility}  

\usepackage{hyperref}

\usepackage{orcidlink}

\usepackage{enumitem}

\usepackage{algorithm}
\usepackage{algpseudocode}

\usepackage{amsmath,amssymb}

\usepackage{wrapfig}   
\usepackage{booktabs}  
\usepackage{multirow}  
\usepackage{graphicx}  
\usepackage{amssymb}   
\usepackage{bm}        

\usepackage{xcolor} 
\definecolor{mygray}{gray}{0.92} 

\usepackage{bbm}

\begin{document}

\title{Difference Feedback: Generating Multimodal Process-Level Supervision for VLM Reinforcement Learning}

\titlerunning{Difference Feedback for VLM Alignment}

\author{Feiding \and
Yongkang Zhang \and
Yuhao Liao \and
Zijian Zeng \and
Chunzheng Zhu \and
Yaozong Zheng \and
Yafei Liu \and
Yeling Peng \and
Youwei Wang \and
Sibo Wang \and
Huiming Yang \and
Linglin Liao \and
Shunzhi Yang}

\authorrunning{Feiding et al.}

\institute{Shenzhen International Graduate School, Tsinghua University, China}

\maketitle


\begin{abstract}
    Vision--language models (VLMs) are increasingly aligned via Group Relative Policy Optimization (GRPO)-style training. However, relying solely on terminal outcome rewards yields sparse credit assignment in multi-step reasoning, weakening the linkage between visual evidence and intermediate steps and often causing unstable optimization and visual hallucinations. We propose \emph{Differential Feedback}, which automatically constructs token/step-level supervision masks by repairing erroneous reasoning trajectories, explicitly marking the key positions that require correction. Without costly large-scale step-by-step human annotations, our method enables process-level visual alignment and can be seamlessly integrated into existing GRPO-like frameworks. Experiments on multimodal reasoning benchmarks including MMMStar and MathVista show an average \textbf{3\%} improvement under matched compute budgets. Our approach offers an effective, low-cost solution for accurate vision--reasoning process alignment.
\end{abstract}

\begin{figure}[t]
    \centering
    \includegraphics[width=\linewidth]{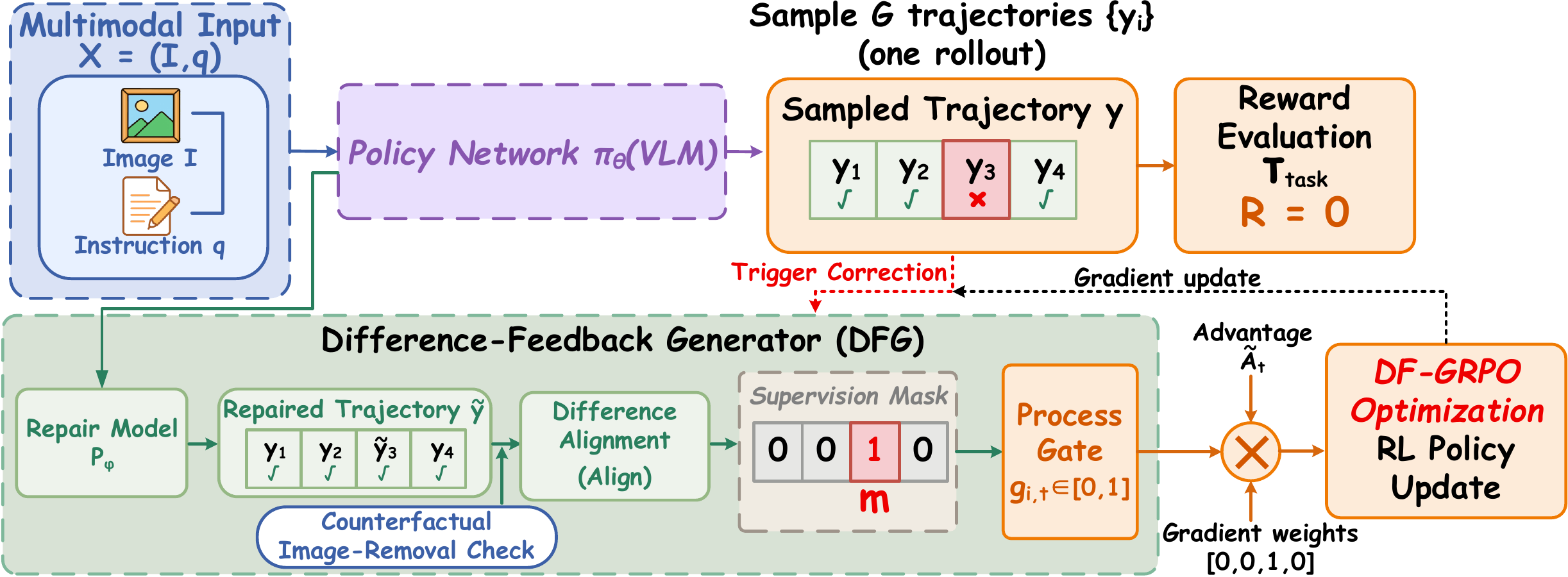}
    \caption{Difference Feedback (DF) provides fine-grained process supervision for VLM alignment. When the policy produces an incorrect trajectory, a small-edit repair is generated; the difference between the two outputs yields a token-level mask that gates gradient updates.
    }
    \label{fig:overview}
\end{figure}

\section{Introduction}
\label{sec:intro}

Vision-Language Models (VLMs) have demonstrated impressive performance on visual
question answering, cross-modal reasoning, and multi-turn dialogue by integrating
visual encoders with large language model generation. To improve instruction
following and alignment, ~\cite{yao2025r1,zha2025vision} introduce
reinforcement learning and preference optimization into VLM training, including
GRPO~\cite{shao2024deepseekmath}, PPO~\cite{schulman2017proximal}, GSPO~\cite{zheng2025group}, and SAPO~\cite{gao2025soft}. Despite impressive engineering progress, the training
signal design universally relies on \emph{outcome supervision}: models receive a
single global reward or preference judgment only after generation completes,
without fine-grained feedback on intermediate reasoning steps or critical visual
decisions.

Outcome supervision creates significant \textbf{credit assignment difficulties}
in multimodal, multi-step reasoning. Different tokens, reasoning steps, and
visual evidence choices contribute unequally to the final outcome; a single
outcome signal amplifies policy gradient variance and makes updates susceptible
to sampling noise—manifesting as training instability, slow convergence, and
wasteful gradient updates. In challenging samples involving fine-grained
recognition, counting, spatial relationships, or cross-modal constraints, errors
typically concentrate in a few critical fragments (\eg, a wrong attribute word, a
misidentified region, or a logical jump). Outcome signals cannot localize these
critical steps, limiting stable improvement on long-tail hard cases. Concrete
failure modes include inability to learn from hard examples and responses that
ignore visual content in favor of language priors.

Process-level supervision is a natural remedy: training a Process Reward Model
(PRM) to score intermediate steps, or combining test-time search with step
scoring. However, explicit process supervision requires expensive step-level
annotation that is difficult to scale; test-time search incurs substantial
computational overhead and is highly sensitive to evaluator quality. In
multimodal tasks, process annotation further requires reliable visual evidence
verification, amplifying cost and noise. Thus, \textbf{obtaining stable,
effective process-level training signals without large-scale annotation or heavy
search overhead} remains a central challenge in VLM alignment.

We propose \textbf{Difference Feedback (DF)} to address this challenge (Fig.~\ref{fig:overview}).
Difference Feedback is not a specific optimizer, but rather a \textbf{general mechanism for generating process-level supervision signals}.
When the policy model produces an incorrect trajectory, we train a multimodal repair model to generate a corrected answer while making relatively small edits to the original response.
We then align the original answer with the repaired answer and automatically derive token-/step-level supervision masks from their differences, explicitly identifying \emph{which positions need to be corrected}.
This supervision signal can be \textbf{plugged into} a variety of alignment objectives, where the difference mask serves as token-level gradient gating to reduce ineffective updates.
As a result, Difference Feedback preserves the scalability of training while providing more stable and interpretable local correction signals for long-horizon multimodal reasoning.

We evaluate our approach on multiple multimodal reasoning and alignment benchmarks.
Experimental results show that Difference Feedback consistently improves performance across different evaluation datasets.
Overall, this work opens a new general pathway for VLM alignment: automatically constructing process-level supervision via output differences, leading to stronger multimodal alignment and reasoning capabilities.

\vspace{2pt}
\noindent\textbf{Contributions.}
The main contributions of this work are summarized as follows:
\begin{itemize}[leftmargin=*,itemsep=2pt,topsep=2pt]
    \item \textbf{We introduce the Difference Feedback (DF) mechanism.}
    Without modifying the neural architecture, the objective function, or requiring large-scale step-by-step human annotations, DF provides a general method for generating process-level supervision.
    Through a ``repair $\rightarrow$ alignment $\rightarrow$ difference mask'' pipeline, the method automatically constructs token-/step-level correction signals, and uses the difference mask to gate token-level updates, reducing the probability of gradients updating irrelevant tokens.

    \item \textbf{Compatibility with diverse alignment objectives.}
    As a plug-in module, Difference Feedback can be integrated into multiple optimization methods such as GRPO, DAPO, GSPO, and PPO, reducing ineffective updates and improving the training efficiency of multimodal models.

    \item \textbf{Consistent improvements on multimodal reasoning and alignment tasks.}
    Across multiple benchmarks, Difference Feedback yields consistent performance gains.
\end{itemize}

\section{Related Work}
\label{sec:related_work}

Alignment and Reinforcement Learning(RL) fine-tuning typically rely on sequence-level outcome
reward/preference signals, causing credit assignment difficulties and
high-variance updates in multi-step generation. Process supervision mitigates
this but requires expensive step-level annotation and PRMs~\cite{lightman2023verify}.
To avoid step-level labeling, one line of work constructs denser token-level
signals from outcome feedback: \cite{chan2024dense} redistributes rewards using
internal reward model information; \cite{li2024r3hf} decomposes contributions for
reward redistribution; \cite{cao2025scar} allocates marginal contributions from a
Shapley perspective. Another line learns token-wise signals from preference
optimization, typified by DPO~\cite{rafailov2023dpo} and token-level
variants~\cite{zhong2024rto,zhou2024treg,chen2025qrm,xia2024inverseq};
\cite{chen2024rlmec} uses per-token generation probability as token-level reward.

Unlike these methods that answer \emph{``how much credit should each token
receive,''} our work asks \emph{``which tokens/steps are the critical bottlenecks
causing errors.''} We localize erroneous positions by introducing lightly-edited
repairs: rather than outputting token-level reward values, we generate token/step
masks via edit alignment (Levenshtein/LCS) and inject them into GRPO/PPO/GSPO
objectives, focusing negative advantage on segments requiring correction. Related
work on minimal contrastive edits~\cite{ross2021mice} demonstrates the value of
minor edits for causal span localization. Self-refinement
methods~\cite{madaan2023selfrefine,kumar2024score} focus on test-time iterative
revision. Implicit process rewards~\cite{yuan2024implicitprm} provide token-level
reward allocation but face a language capability--reasoning trade-off that limits
scalability. To our knowledge, we are the first to introduce process-level
supervision into multimodal RL training, using repair differences to improve
multimodal alignment under sparse outcome supervision.

\section{Method}
\label{sec:method}

\subsection{Notation and Problem Setup}
\label{sec:setup}

We are given a multimodal input $x=(\mathbf{I},\mathbf{q})$, where $\mathbf{I}$ denotes an image (or a sequence of video frames) and $\mathbf{q}$ denotes a textual instruction/question.
The policy model (VLM) is denoted by $\pi_{\theta}$, which autoregressively generates an output sequence
\[
\mathbf{y}=(y_1,\ldots,y_T).
\]
We define the state at step $t$ as $s_t=(x,\mathbf{y}_{<t})$.
We consider a family of alignment optimization objectives $\mathcal{J}(\theta)$, whose training signals are typically derived from \emph{terminal rewards/preferences}, e.g., a sequence-level reward $R(x,\mathbf{y})$ or a preference pair $(\mathbf{y}^+,\mathbf{y}^-)$.
Such terminal supervision is sparse, making credit assignment difficult; in practice, this manifests as high-variance gradients and many ineffective updates.
To address this issue, we propose \textbf{Difference Feedback (DF)}: a general mechanism for constructing process-level supervision signals, which can be plugged into and enhance various optimization objectives such as GRPO, PPO, and GSPO.

\subsection{Difference Feedback Generation (DF)}
\label{sec:dfg}

\paragraph{Core idea.}
When the policy output $\mathbf{y}$ is incorrect (or receives a low score), we use a \textbf{multimodal repair model} $\rho_{\phi}$ to produce a repaired output $\tilde{\mathbf{y}}$ with \textbf{minimal edits}.
We then align $\mathbf{y}$ and $\tilde{\mathbf{y}}$ and derive a token-/step-level supervision mask $\mathbf{m}$ from their differences, indicating \emph{which positions should be updated}.
We first train the multimodal repair model (described in Sec.~\ref{sec:repair_train}), and then use it to generate difference feedback (detailed in Sec.~\ref{sec:repair_model}).

\subsection{Train the Repair Model: Two-Stage (Supervised Fine-Tuning(SFT)\,+\,RL)}
\label{sec:repair_train}

\subsubsection{Stage 1: SFT Warm Start}
\label{sec:repair_sft}

We manually construct a small repair dataset (1,000 samples)
$\mathcal{D}_{\text{rep}}=\{(x,\mathbf{y},\mathbf{a},\mathbf{y}^*)\}$,
where $\mathbf{y}$ denotes an incorrect output, $\mathbf{y}^*$ denotes the repaired target output, and $\mathbf{a}$ denotes the correct reference answer.
The pair $(\mathbf{y},\mathbf{a})$ can be obtained by prompting the policy model to generate candidate outputs followed by correctness filtering, while $\mathbf{a}$ and $\mathbf{y}^*$ are manually annotated.

We initialize the repair model using supervised fine-tuning (SFT) with the objective:
\begin{equation}
    \mathcal{L}_{\text{SFT}}(\phi)
    =
    - \mathbb{E}_{(x,\mathbf{y},\mathbf{a},\mathbf{y}^*)\sim \mathcal{D}_{\text{rep}}}
    \Bigg[
    \sum_{t=1}^{T^*}
    \log \rho_{\phi}(y^*_t\mid x,\mathbf{y},\mathbf{a},\mathbf{y}^*_{<t})
    \Bigg].
\end{equation}

\subsubsection{Stage 2: Learning to Repair with Reinforcement Learning}
\label{sec:repair_rl}

To prevent the repair model from rewriting the entire response, we further train it with reinforcement learning to encourage
\textbf{correct repairs with relatively small edits}.

For a repaired output $\tilde{\mathbf{y}}$, we define a task correctness reward
$r_{\text{task}}(x,\tilde{\mathbf{y}})\in[0,1]$.
This reward reuses the correctness evaluation used for training the policy model:
for mathematical problems we use a rule-based reward that checks only the final answer,
while for general tasks we use an LLM-based judge.

We also define an edit cost measuring how many tokens in $\mathbf{y}$ are modified:
\begin{equation}
    \Delta_{\text{edit}}(\mathbf{y},\tilde{\mathbf{y}})
    =
    \mathrm{EditDist}(\mathbf{y},\tilde{\mathbf{y}}).
\end{equation}

\paragraph{Reward-hacking auditing (gating).}
We introduce an auditor based on Qwen2.5-VL-72B-Instruct to inspect repair trajectories.
The audit outcome is formalized as a gating factor
$g_{\text{aud}}(x,\mathbf{y},\tilde{\mathbf{y}})\in\{0,1\}$.

If the auditor detects potential reward-hacking behaviors
(e.g., the answer remains correct after removing the image while the reasoning still claims visual evidence,
template-based shortcuts, or spurious difference masks caused by compensatory reasoning or paraphrasing—i.e., mask hallucination),
we set $g_{\text{aud}}=0$; otherwise $g_{\text{aud}}=1$.

The auditor does not participate in gradient updates and is used only for sample filtering and failure-mode statistics.
We report both the audit hit rate and the agreement rate with manual inspection to evaluate its reliability.

\paragraph{Image-removal penalty (C1).}
To discourage shortcut behaviors where the model can answer correctly without using visual information,
we construct a counterfactual sample without visual input
$x_{\varnothing}=(\mathbf{I}_{\varnothing},\mathbf{q})$,
where $\mathbf{I}_{\varnothing}$ is a blank image (all zeros) or a random-noise image.

We define a visual-dependence gating factor:
\begin{equation}
    g_{\text{vis}}(x,\tilde{\mathbf{y}})
    =
    \mathbb{I}\!\left[
    r_{\text{task}}(x_{\varnothing},\tilde{\mathbf{y}})<1
    \right]
    \in\{0,1\}.
\end{equation}

If the repaired output is still judged correct after removing the image,
we treat it as not relying on genuine visual evidence and set the reward to zero ($g_{\text{vis}}=0$).
We further verify with the policy model whether the problem can be solved using text alone;
if so, the corresponding training sample is removed.

\paragraph{Final repair reward (with auditing and C1).}
We integrate the auditing and visual-dependence constraints into the repair reward:
\begin{equation}
    r_{\text{rep}}(x,\mathbf y,\tilde{\mathbf y})
    =
    g_{\text{aud}}(x,\mathbf y,\tilde{\mathbf y})
    \cdot
    g_{\text{vis}}(x,\tilde{\mathbf y})
    \cdot
    \left(
    r_{\text{task}}(x,\tilde{\mathbf y})
    -
    \lambda \cdot
    \frac{\Delta_{\text{edit}}(\mathbf y,\tilde{\mathbf y})}{\max(1,|\mathbf y|)}
    \right).
    \label{eq:r_rep_gate}
\end{equation}


In Stage 2, the repair model is optimized using the same reinforcement learning algorithm as the policy (for example, if the policy uses GRPO, the repair model also uses GRPO), with the only difference being the reward function.

\subsection{Obtaining Difference Feedback}
\label{sec:repair_model}

Given incorrect trajectory $\mathbf{y}$, we independently sample $n$ repairs from
the repair model:
\begin{equation}
    \tilde{\mathbf{y}}^{(i)} \sim
    \rho_{\phi}(\cdot \mid x,\mathbf{y},\mathbf{a}),
    \quad i=1,\dots,n,
\end{equation}
where $\mathbf{a}$ is a reference signal (annotated answer, a preferred response
in preference data, or a correct trajectory uniformly sampled from
$\mathcal{G}_{\text{cor}}$). The repair model is initialized as
$\phi \leftarrow \theta$ to keep repairs in-distribution. We define the feasible
set:
\begin{equation}
    \mathcal{S}(x,\mathbf{y})
    =
    \left\{
    \tilde{\mathbf{y}}^{(i)}
    \mid
    r_{\text{task}}(x,\tilde{\mathbf{y}}^{(i)}) = 1
    \right\},
\end{equation}
and select the optimal (minor-edit) repair:
\begin{equation}
    \tilde{\mathbf{y}}
    =
    \arg\min_{\tilde{\mathbf{y}} \in \mathcal{S}(x,\mathbf{y})}
    \Delta_{\text{edit}}(\mathbf{y},\tilde{\mathbf{y}}).
    \label{eq:rs_optimal}
\end{equation}

\subsubsection{Difference Alignment and Supervision Mask.}
\label{sec:align_mask}

Given original output $\mathbf{y}$ and repair $\tilde{\mathbf{y}}$, the alignment
operator $\mathcal{A}$ yields two-sided token-level masks:
\begin{equation}
    (\mathbf{m},\mathbf{m}')=\mathcal{A}(\mathbf{y},\tilde{\mathbf{y}}),
    \label{eq:mask_trunc}
\end{equation}
where $\mathbf{m}\in\{0,1\}^{T}$ marks tokens in $\mathbf{y}$ requiring
correction and $\mathbf{m}'\in\{0,1\}^{\tilde{T}}$ marks the corresponding
repair tokens in $\tilde{\mathbf{y}}$. Concretely, $\mathcal{A}$ is implemented
via the Levenshtein edit path (or LCS): $m_t=1$ at substitution/deletion/insertion
positions and $m_t=0$ otherwise.

\paragraph{Soft Weights (Optional).}
Beyond binary masks, soft weights $w_t\in[0,1]$ with distance-based decay can be
used:
\begin{equation}
    w_t = m_t \cdot \exp\!\big(-\alpha (t-\tau)\big).
    \label{eq:soft_weight}
\end{equation}
For clarity we use binary masks below; soft weights are a direct drop-in replacement.

\subsection{Plug-in Injection of Difference Feedback: A Unified View}
\label{sec:plugin_view}

We formalize the \emph{plug-in} nature of Difference Feedback (DF) as follows.
For any objective based on token-level log-likelihood, DF introduces a gating term $g_t$—either a binary mask or a soft weight—to reweight the contribution of each token to the update:

\begin{equation}
    g_t \in [0,1],\qquad
    g_t = m_t \ \ \text{or}\ \ g_t=w_t.
    \label{eq:gating}
\end{equation}

Intuitively, $g_t$ determines whether a token should receive credit assignment from terminal supervision.
Tokens identified by DF as responsible for the error receive higher update weights, while irrelevant tokens are suppressed.

Due to space limitations, we present below the concrete formulations of DF when integrated with GRPO and PPO.

\subsection{DF-GRPO: Repair-Pair-Based Group Relative Policy Optimization}
\label{sec:df_grpo}

\paragraph{Sampling an initial group $\mathcal{G}'$.}
For each input $x$, we sample a group of candidate trajectories from the old policy $\pi_{\theta_{\text{old}}}$:
\begin{equation}
    \mathcal{G}'(x)=\{y_i\}_{i=1}^{G'},\qquad
    y_i \sim \pi_{\theta_{\text{old}}}(\cdot\mid x),
    \label{eq:Gprime_sampling}
\end{equation}
where $|\mathcal{G}'(x)|=G'$.

\paragraph{Splitting trajectories by reward threshold $\tau_r$.}
Using the terminal reward $r(x,y)$ and a threshold $\tau_r$, we partition $\mathcal{G}'(x)$ into correct and erroneous trajectories:
\begin{equation}
    \mathcal{G}_{\text{cor}}(x)=\{y\in \mathcal{G}'(x)\mid r(x,y)\ge \tau_r\},\qquad
    \mathcal{G}_{\text{err}}(x)=\{y\in \mathcal{G}'(x)\mid r(x,y)< \tau_r\}.
    \label{eq:split_Gprime}
\end{equation}

\paragraph{Sampling a reference $a$ from correct trajectories.}
If $\mathcal{G}_{\text{cor}}(x)\neq\emptyset$, we uniformly sample a reference trajectory
\begin{equation}
    a \sim \mathrm{Unif}\big(\mathcal{G}_{\text{cor}}(x)\big).
    \label{eq:sample_a_from_correct}
\end{equation}

If no correct trajectory exists, this sample cannot provide a meaningful reward signal and is skipped.

\paragraph{Repairing erroneous trajectories to obtain $\mathcal{G}_{\text{rep}}$.}
For each erroneous trajectory $y_i\in \mathcal{G}_{\text{err}}(x)$, we generate a repaired trajectory using the repair model:
\begin{equation}
    \tilde{y}_i \sim \rho_{\phi}(\cdot \mid x, y_i, a),
    \qquad y_i\in \mathcal{G}_{\text{err}}(x),
    \label{eq:repair_each_error}
\end{equation}
and collect them into
\begin{equation}
    \mathcal{G}_{\text{rep}}(x)=\{\tilde{y}_i \mid y_i\in \mathcal{G}_{\text{err}}(x)\}.
    \label{eq:Grep_def}
\end{equation}

\paragraph{Constructing the optimization group $\mathcal{G}$ (size $G$).}
We remove all correct trajectories from $\mathcal{G}'(x)$ and add the repaired versions of erroneous trajectories:
\begin{equation}
    \mathcal{G}(x)
    =
    \big(\mathcal{G}'(x)\setminus \mathcal{G}_{\text{cor}}(x)\big)
    \ \cup\
    \mathcal{G}_{\text{rep}}(x)
    =
    \mathcal{G}_{\text{err}}(x)\ \cup\ \mathcal{G}_{\text{rep}}(x),
    \label{eq:new_group_G}
\end{equation}
and denote
\begin{equation}
    |\mathcal{G}(x)| = G.
    \label{eq:G_def}
\end{equation}

\paragraph{Group-normalized advantages in GRPO.}
For each trajectory $z\in\mathcal{G}(x)$, we compute a sequence-level advantage $\widehat{A}(z)$ obtained by normalizing rewards within the group $\mathcal{G}(x)$.
The token-level likelihood ratio is defined as
\begin{equation}
    w_{z,t}(\theta)
    =
    \frac{\pi_{\theta}(z_t\mid x, z_{<t})}
    {\pi_{\theta_{\text{old}}}(z_t\mid x, z_{<t})}.
    \label{eq:ratio_z_t}
\end{equation}

\paragraph{Two-sided difference gating for error--repair pairs.}
For each error--repair pair $(y_i,\tilde{y}_i)$, we apply an alignment operator $\mathcal{A}$ (see Sec.~\ref{sec:align_mask}) to obtain token-level difference masks on both sides:
\begin{equation}
    \big(m_{i,1:|y_i|},\ m'_{i,1:|\tilde{y}_i|}\big)
    =
    \mathcal{A}(y_i,\tilde{y}_i),
    \label{eq:mask_pair_two_sided}
\end{equation}
where $m_{i,t}\in\{0,1\}$ marks positions in the erroneous trajectory $y_i$ that require correction, and $m'_{i,t}\in\{0,1\}$ marks the corresponding repair tokens in $\tilde{y}_i$.

We apply the gating to the two trajectories respectively:
\begin{equation}
    g_{z,t}(x)=
    \begin{cases}
        m_{i,t},  & z = y_i,\ \ y_i\in \mathcal{G}_{\text{err}}(x),\\[3pt]
        m'_{i,t}, & z = \tilde{y}_i,\ \ \tilde{y}_i\in \mathcal{G}_{\text{rep}}(x).
    \end{cases}
    \label{eq:pair_two_sided_gating}
\end{equation}

The resulting token-level gated advantage is
\begin{equation}
    \widehat{A}^{\text{DF}}_{z,t} = g_{z,t}\cdot \widehat{A}(z).
    \label{eq:df_adv_pair}
\end{equation}

\paragraph{DF-GRPO objective.}
\begin{align}
    \mathcal{J}_\text{DF-GRPO}(\theta)
    &=
    \mathbb{E}_{x \sim \mathcal{D},\, \mathcal{G}(x)}
    \Bigg[
    \frac{1}{G}
    \sum_{z\in\mathcal{G}(x)}
    \frac{1}{|z|}
    \sum_{t=1}^{|z|}
    \mathcal{L}_{z,t}^{\text{clip}}
    \Bigg]
    -
    \beta_{\text{kl}}\,\mathcal{K}(\theta),
    \label{equ:df_grpo}
\end{align}
where
\begin{equation}
    \mathcal{L}_{z,t}^{\text{clip}}
    =
    \min\Big(
    w_{z,t}(\theta)\,\widehat{A}^{\text{DF}}_{z,t},
    \mathrm{clip}(w_{z,t}(\theta),1-\epsilon,1+\epsilon)\,\widehat{A}^{\text{DF}}_{z,t}
    \Big).
    \label{eq:clip_loss_pair}
\end{equation}

\subsection{DF-PPO: Masked PPO-Clip (General Advantage Injection)}
\label{sec:df_ppo}

PPO defines the token-level likelihood ratio:
\begin{equation}
    r_t(\theta)=\frac{\pi_{\theta}(y_t\mid s_t)}{\pi_{\theta_{\text{old}}}(y_t\mid s_t)}.
\end{equation}

The standard PPO-Clip surrogate objective (to maximize) is
\begin{equation}
    \mathcal{J}_{\text{PPO}}(\theta)=
    \mathbb{E}\Bigg[
    \sum_{t=1}^{T}
    \min\Big(
    r_t(\theta)\hat{A}_t,\
    \mathrm{clip}(r_t(\theta),1-\epsilon,1+\epsilon)\hat{A}_t
    \Big)
    \Bigg].
\end{equation}

\paragraph{Key idea: different gating for correct and incorrect trajectories.}
Difference Feedback is applied only to ``incorrect/low-reward'' trajectories in order to localize the segments that require correction.
For ``correct/high-reward'' trajectories, we retain the full PPO positive reinforcement signal to avoid discarding useful learning signals.

Specifically, let the terminal reward (or correctness signal) be $R(x,\mathbf{y})$ and the threshold be $\tau_r$.
We define the gating coefficient as
\begin{equation}
    g_t(x,\mathbf{y})=
    \begin{cases}
        1, & R(x,\mathbf{y}) \ge \tau_r,\\
        m_t(\mathbf{y},\tilde{\mathbf{y}}), & R(x,\mathbf{y}) < \tau_r,
    \end{cases}
    \label{eq:gating_correct_incorrect}
\end{equation}
where, when $R(x,\mathbf{y}) < \tau_r$, we generate a repaired trajectory
$\tilde{\mathbf{y}}\sim\rho_{\phi}(\cdot\mid x,\mathbf{y},\mathbf{a})$
using the repair model,
and obtain the truncated difference mask $m_t$ via alignment (Eq.~\eqref{eq:mask_trunc}).

\paragraph{Masked advantage.}
We gate the advantage as
\begin{equation}
    \hat{A}^{\text{DF}}_t = g_t \cdot \hat{A}_t,
    \label{eq:df_adv}
\end{equation}
which yields the DF-PPO objective (to maximize):
\begin{equation}
    \mathcal{J}_{\text{DF-PPO}}(\theta)=
    \mathbb{E}\Bigg[
    \sum_{t=1}^{T}
    \min\Big(
    r_t(\theta)\hat{A}^{\text{DF}}_t,\
    \mathrm{clip}(r_t(\theta),1-\epsilon,1+\epsilon)\hat{A}^{\text{DF}}_t
    \Big)
    \Bigg]
    -\beta_{\text{kl}}\mathcal{K}(\theta).
\end{equation}

When the trajectory is correct, $g_t\equiv 1$ and DF-PPO reduces to the standard PPO update.
When the trajectory is incorrect, $g_t=m_t$ restricts negative advantages to only the tokens or steps that require correction, thereby significantly reducing erroneous credit assignment and ineffective updates caused by sparse terminal supervision.

\subsection{Why Difference Feedback Reduces Ineffective Updates: A Counterfactual Path Perspective (Brief)}
\label{sec:variance}

The central role of Difference Feedback (DF) is not to claim a strict variance-reduction guarantee for arbitrary tasks, but rather to mitigate misattribution under terminal supervision by leveraging a \textbf{counterfactual reference trajectory} provided by the repair model.

In this work, the quality of a trajectory is \textbf{defined in terms of terminal correctness}.
Given a reward function $r_{\text{task}}$, any generated sequence that is judged correct at the final step can be regarded as an \emph{effective trajectory} and can therefore serve as a valid training signal.
We do not require the intermediate reasoning in the repaired trajectory to match the original trajectory or to follow a unique reasoning form.
Instead, the key question is whether there exists a trajectory that can flip the terminal judgment from incorrect to correct, and \emph{which local segments must be modified to achieve such a flip}.

Concretely, when the policy output $\mathbf{y}$ is an incorrect or low-scoring trajectory, we construct a repaired output $\tilde{\mathbf{y}}$ that satisfies the correctness criterion while applying \emph{relatively small edits}, as described in Eq.~\eqref{eq:rs_optimal}.
We then apply an alignment operator to derive a difference mask and its truncated variant (Eq.~\eqref{eq:mask_trunc}).
This construction implies a direct and interpretable property:
\textbf{there exists a counterfactual path that flips the terminal judgment from incorrect to correct, and the required modifications are concentrated in the difference segments}.
The small-edit constraint further discourages opportunistic success via large-scale rewriting, making the resulting difference mask more likely to highlight the key bottleneck segments responsible for the judgment flip.
Consequently, the difference mask can be viewed as a local correction indicator that is \emph{sufficient to flip the terminal decision}.

Under purely terminal reward or preference supervision, sequence-level advantages are typically broadcast across the entire generated output.
This causes negative attribution to accumulate on many tokens that are unrelated to the actual error, resulting in substantial ineffective updates and unstable training dynamics.
DF addresses this issue by applying token-level gating $g_t$ (Eq.~\eqref{eq:gating}) on erroneous trajectories, so that negative advantages and local preference constraints \textbf{primarily act on the difference segments indicated by the counterfactual path}, rather than uniformly affecting the entire sequence.

It is important to emphasize that a counterfactual modification being \emph{sufficient to flip the judgment} does \emph{not} imply that masked tokens are causally irrelevant to the error.
Compensatory effects, solution multiplicity, or semantic rewriting may cause the apparent small edits to deviate from the true causal bottleneck.
However, the goal of the gating mechanism is not strict causal attribution.
Instead, it aims to \textbf{reduce the broadcast of negative signals and the update noise on irrelevant tokens} under sparse terminal supervision, thereby improving learning efficiency and training stability.

\subsection{Computational Complexity and Scalability}
\label{sec:complexity}

The additional computational overhead introduced by Difference Feedback mainly arises from a single inference pass of the repair model, which is triggered only for incorrect samples. The expected cost can be approximated as:
\begin{equation}
    \mathrm{Cost}_{\text{DF}}
    \approx
    \mathrm{Cost}_{\text{base}}
    +
    p_{\text{err}} \cdot \mathrm{Cost}\big(\rho_{\phi}\text{ decode}\big),
\end{equation}
where $p_{\text{err}}$ denotes the proportion of samples that trigger the repair process, which typically decreases as training progresses.

Unlike test-time search methods, Difference Feedback does not introduce exponential branching during decoding. Consequently, it maintains favorable scalability and can be efficiently applied to large-scale VLM alignment training.

\section{Experiments}
\setlength{\tabcolsep}{3pt}

\subsection{Implementation Details}
\label{subsec:Implementation Details}

We conduct experiments using the base models \textbf{Qwen2.5-VL-7B} and \textbf{Qwen2.5-VL-32B}~\cite{bai2025qwen2.5vl}.
The standard baseline is trained with GSPO, while our method uses DF-GSPO.
Training is performed on our curated multimodal mathematical reasoning dataset (image–text pairs) using 32 NVIDIA A800 GPUs.

Unless otherwise specified, we use the following hyperparameters:
a rollout batch size of 128, rollout temperature of 0.7, and a learning rate of $1\times10^{-6}$.
The reward threshold $\tau_r$ is set to 1, $n$ is set to 2 and the edit penalty coefficient $\lambda$ is set to 0.5.

For the base model \textbf{InternVL3.5-8B-MPO}~\cite{wang2025internvl3_5}, we compare GRPO with DF-GRPO.
Training is conducted using the MMPR-Tiny dataset~\cite{wang2025internvl3_5}.
The rollout batch size is 512, rollout temperature is 0.7, and the learning rate is $1\times10^{-6}$.
We use the same reward threshold $\tau_r=1$ , $n$ is set to 2 and edit penalty coefficient $\lambda=0.5$.

\subsubsection{Compute-Matching Protocol}
\label{sec:appendix-compute}

To ensure fair comparison, we use identical model architectures and optimization settings across methods,
and match the overall training computation by accounting for the \textbf{total GPU time} consumed.

Specifically, for DF-based methods we include both the cost of training the repair model and the cost of policy optimization.
For example, the GPU time used by the GRPO baseline is matched to the sum of
(i) the GPU time spent training the repair model and
(ii) the GPU time spent training the policy model with DF-GRPO.
This protocol ensures that all compared methods operate under comparable computational budgets.

\subsection{Main Results}
\label{sec:exp_result}

Table~\ref{tab:main_result} presents comprehensive results at 7B and 32B scales
across benchmarks covering both specialized and general multimodal reasoning.

\begin{table}[t]
    \caption{\textbf{Main experimental results.} For selected experiments, we report the mean over five random seeds together with the 95\% bootstrap confidence interval (mean $\pm$ 95\% CI). Improvements over the baseline methods are statistically significant under a paired bootstrap test ($p < 0.01$).}
    \vskip 0.1in
    \centering
    \scalebox{0.7}{
        \setlength{\tabcolsep}{3pt}
        \begin{tabular}{lccccccc}
            \toprule
            \textbf{Model}  & \textbf{MathVista} & \textbf{MMStar} & \textbf{MMMU} & \textbf{MathVerse} & \textbf{MathVision} & \textbf{AI2D}   \\
            \midrule
            InternVL3.5-8B-MPO \cite{wang2025internvl3_5}      & 75.9 & --   & 71.2 & 54.8 & 52.6 & --   \\
            InternVL3.5-8B-MPO (PRIME \cite{cui2025process})   & 78.5 & --   & 73.6 & 61.9 & 56.9 & --   \\
            InternVL3.5-8B-MPO (GRPO)                          & 78.4 & --   & 73.4 & 61.5 & 56.8 & --   \\
            \rowcolor{mygray}
            InternVL3.5-8B-MPO (\textbf{DF-GRPO})             & $79.9_{\pm1.1}$ & --   & $74.8_{\pm0.6}$ & $63.5_{\pm0.9}$ & $58.2_{\pm1.0}$ & --   \\
            \midrule
            Qwen2.5-VL-7B \cite{qwen2_5_vl}                   & 68.2 & 63.9 & 58.6 & 49.2 & 25.1 & 83.9 \\
            ThinkLite-7B \cite{thinklite-vl}                   & 74.3 & 63.7 & 53.1 & 52.2 & 29.9 & 83.0 \\
            R1-ShareVL-7B \cite{yao2025r1}                     & 75.4 & 67.0 & 58.1 & 52.8 & 29.5 & 84.5 \\
            Vision-G1 \cite{zha2025vision}                     & 76.1 & 66.0 & 53.4 & --   & --   & --   \\
            Vision-R1 \cite{huang2025vision}                   & 73.5 & --   & --   & 52.4 & --   & --   \\
            Qwen2.5-VL-7B (GSPO)                               & $73.6_{\pm1.0}$ & $65.9_{\pm0.8}$ & $56.7_{\pm1.5}$ & $50.8_{\pm1.3}$ & $27.4_{\pm1.4}$ & $84.5_{\pm0.8}$ \\
            \rowcolor{mygray}
            Qwen2.5-VL-7B (\textbf{DF-GSPO})                  & $76.1_{\pm1.8}$ & $68.3_{\pm1.3}$ & $58.4_{\pm0.8}$ & $53.7_{\pm0.9}$ & $30.4_{\pm1.2}$ & $85.6_{\pm0.6}$ \\
            \midrule
            Qwen2.5-VL-32B \cite{qwen2_5_vl}                  & 74.7 & 69.5 & 70.0 & 49.9 & 38.4 & 84.6 \\
            R1-ShareVL-32B \cite{yao2025r1}                    & 77.6 & 70.2 & 70.1 & 59.0 & 40.3 & 86.2 \\
            Qwen2.5-VL-32B (GSPO)                              & $75.5_{\pm1.3}$ & $68.3_{\pm1.4}$ & $66.8_{\pm0.9}$ & $57.2_{\pm0.7}$ & $35.7_{\pm1.2}$ & $85.7_{\pm0.5}$ \\
            \rowcolor{mygray}
            Qwen2.5-VL-32B (\textbf{DF-GSPO})                 & $78.8_{\pm1.2}$ & $71.5_{\pm1.5}$ & $70.6_{\pm0.8}$ & $61.3_{\pm1.8}$ & $41.6_{\pm0.9}$ & $87.8_{\pm0.7}$ \\
            \bottomrule
        \end{tabular}
    }
    \label{tab:main_result}
    \vskip -0.2in
\end{table}

\noindent\textbf{Comparison with RL-trained MLLMs.}
Training Qwen2.5-VL with GSPO and DF-GSPO under matched GPU time, DF-GSPO
consistently outperforms GSPO at 7B—with notable gains on MathVista and
MathVerse. GSPO exhibits performance drops on MMMU, likely due to catastrophic
forgetting from wasteful updates; DF-GSPO substantially mitigates this. At 32B,
standard GSPO shows drops on multiple metrics, suggesting larger models are more
susceptible to wasteful updates; DF-GSPO recovers and surpasses all baselines.
Using InternVL3.5-8B-MPO as a distinct backbone, DF-GRPO outperforms GRPO,
confirming the generality of the DF mechanism.

\noindent\textbf{Comparison with Other Process Reward Methods.}
PRIME~\cite{cui2025process} applies implicit process rewards~\cite{yuan2024implicitprm}.
DF outperforms PRIME across all evaluated metrics, likely because PRIME suffers
from a fundamental objective inconsistency: PRM training progressively degrades
the model's base language capability, limiting scalability.

\subsection{Ablation Study}
\label{sec:exp_ablation}

\paragraph{Goal and Setup.}
To quantify the contribution of each component in Difference Feedback (DF), we conduct an ablation study on \texttt{Qwen2.5-VL-7B} using the same training data, hyperparameters, and evaluation protocol as in the main experiments.
Unless explicitly removed, all other components remain identical: the same GSPO configuration, rollout temperature and batch size, KL coefficient, and number of training steps.

\paragraph{Control Variants.}
\vspace{2pt}
\noindent The ablated variants are listed below (the default configuration is denoted as \textbf{DF-GSPO}):
\begin{itemize}[leftmargin=*,itemsep=2pt,topsep=2pt]
    \item \textbf{GSPO (w/o DF):} No difference-feedback gating is applied, equivalent to standard GSPO ($g_{t}\equiv 1$).

    \item \textbf{DF w/ LCS Align:} The alignment operator $\mathcal{A}$ is replaced from the Levenshtein shortest-edit-path alignment to LCS-based alignment.

    \item \textbf{DF w/ Soft Weight:} Replace the binary mask with the soft weighting scheme $w_t$ defined in Eq.~\eqref{eq:soft_weight}.

    \item \textbf{DF w/ SFT-only Repair:} The repair model is trained only with Stage-1 repair SFT, without the Stage-2 RL stage that encourages correct repairs with relatively small edits.

    \item \textbf{DF w/o Edit Penalty:} Remove the edit penalty term in the repair reward (Eq.~\eqref{eq:r_rep_gate}, setting $\lambda=0$) to examine the effect of excessive rewriting on mask quality.

    \item \textbf{DF w/o C1 (Image-Removal Penalty):} Remove the image-removal penalty (Eq.~\eqref{eq:r_rep_gate}), i.e., the counterfactual test on $x_{\varnothing}$ is disabled. This variant evaluates the contribution of C1 in suppressing degeneration caused by text-only shortcuts or ignoring visual evidence.
\end{itemize}

\begin{table}[!htbp]
    \caption{\textbf{Ablation study (7B).} Impact of each component on multimodal
    reasoning and alignment performance. Accuracy (\%), higher is better.}
    \vskip 0.08in
    \centering
    \scalebox{0.72}{
        \setlength{\tabcolsep}{3.2pt}
        \begin{tabular}{lccccccc}
            \toprule
            \textbf{Method}  & \textbf{MathVista} & \textbf{MMStar} & \textbf{MMMU} & \textbf{MathVerse} & \textbf{MathVision} & \textbf{AI2D}   \\
            \midrule
            GSPO (w/o DF)                    & 73.6 & 65.9 & 56.7 & 50.8 & 27.4 & 84.5 \\
            DF w/ LCS Align                     & 75.9 & 67.5 & 57.1 & 52.3 & 29.7 & 84.6   \\
            DF w/ Soft Weight                   & 75.5 & 67.6 & 57.2 & 53.5 & 30.3 & 85.4   \\
            DF w/ SFT-only Repair               & 73.3 & 65.1 & 56.5 & 50.5 & 26.2 & 84.3   \\
            DF w/o Min-Edit ($\lambda=0$)       & 75.4 & 67.0 & 56.9 & 52.1 & 28.1 & 84.8   \\
            DF w/o C1 (Image-Removal)           & 75.2 & 68.1 & 57.0 & 53.3 & 30.2 & 84.9     \\
            \midrule
            \rowcolor{mygray}
            \textbf{DF-GSPO (Full)}          & 76.1 & 68.3 & 58.4 & 53.7 & 30.4 & 85.6 \\
            \bottomrule
        \end{tabular}
    }
    \label{tab:ablation_7b}
    \vskip -0.18in
\end{table}

\paragraph{Result analysis.}
(1) Removing DF (GSPO w/o DF) leads to a clear overall drop, indicating that difference-based gating effectively alleviates misattribution under terminal supervision.\\
(2) LCS alignment performs worse than the shortest edit path, as it does not model \emph{substitution} operations well, resulting in inaccurate difference localization and degraded performance.\\
(3) The SFT-only repair model degrades substantially, suggesting that without Stage-2 (SFT+RL) training it cannot reliably produce counterfactual answers that are both correct and involve relatively small edits, which in turn reduces the quality of the difference masks.\\
(4) DF w/o Edit Penalty ($\lambda=0$) consistently performs worse, indicating that excessive rewriting expands the difference mask from critical corrections to many irrelevant edits, thereby increasing gating noise.\\
(5) Removing C1 leads to a mild degradation, suggesting that the counterfactual image-removal test helps suppress shortcut behaviors such as ignoring visual inputs or relying on text-only heuristics.

\paragraph{Sensitivity analysis of $\lambda$.}
We analyze the effect of the edit-penalty coefficient $\lambda$ (Eq.~\eqref{eq:r_rep_gate}) on the potential mismatch between superficial edits and true causal errors, which may lead to reward hacking.
Repair accuracy is evaluated using the auditing model together with manual verification.

As shown in Table~\ref{tab:lambda_repair_acc_wide_5pts}, when $\lambda<0.5$, the repair accuracy remains relatively stable.
As $\lambda$ increases further, the repair model becomes increasingly reluctant to modify the original output; consequently, many errors cannot be corrected to the correct answer, and reward-hacking behaviors become more likely.

\paragraph{Effect of the number of repair samples $n$.}
In Table~\ref{tab:ablation_n}, we study the performance of DF-GSPO under different numbers of repair samples $n$ while keeping the total GPU time fixed. We observe that when $n=2$, the model trained with DF-GSPO achieves the best performance on MathVista.
As the number of samples increases, the additional computational overhead reduces the effective number of optimization steps within the same GPU budget, resulting in smaller performance gains.

\begin{table}[!htbp]
    \centering
    \begin{minipage}[t]{0.44\textwidth}
        \centering
        \caption{\textbf{Sensitivity of $\lambda$.} Acc@1: pass@1; Acc@2: pass@2 ($n=2$).}
        \vskip 0.08in
        \scalebox{0.88}{
            \setlength{\tabcolsep}{5pt}
            \begin{tabular}{lccccc}
                \toprule
                $\lambda$   & 0.2  & 0.4  & 0.5  & 0.6  & 0.8  \\
                \midrule
                Acc@1 (\%)  & 81.3 & 81.1 & 80.9 & 78.2 & 74.2 \\
                Acc@2 (\%)  & 83.2 & 82.9 & 82.1 & 79.6 & 75.7 \\
                \bottomrule
            \end{tabular}
        }
        \vskip -0.1in
        \label{tab:lambda_repair_acc_wide_5pts}
    \end{minipage}%
    \hspace{0.08\textwidth}
    \begin{minipage}[t]{0.43\textwidth}
        \centering
        \caption{\textbf{Effect of sampling count $n$ (matched GPU time).}}
        \vskip 0.08in
        \scalebox{0.88}{
            \setlength{\tabcolsep}{5pt}
            \begin{tabular}{lcccc}
                \toprule
                $n$         & 1     & 2     & 3     & 4     \\
                \midrule
                Acc@1 (\%)  & 75.52 & 76.11 & 76.07 & 75.95 \\
                \bottomrule
            \end{tabular}
        }
        \vskip -0.2in
        \label{tab:ablation_n}
    \end{minipage}
\end{table}

\section{Conclusion}
\label{sec:conclusion}

We address the widely observed problem of \emph{sparse terminal supervision} in RL-based alignment for VLMs (e.g., GRPO), which leads to difficult credit assignment, high-variance updates, unstable training, slow learning on hard examples, and insufficient utilization of visual evidence.
We propose \textbf{Difference Feedback (DF)}, a novel process-supervision paradigm that automatically constructs token-/step-level supervision through a \textbf{repair $\rightarrow$ alignment $\rightarrow$ masking} pipeline.
Specifically, a multimodal repair model generates a correct version of the erroneous trajectory with relatively small edits; the resulting difference mask is then injected into the alignment objective as a gradient gating signal, focusing negative credit assignment on the critical segments responsible for the error.
This mechanism substantially reduces ineffective updates and yields consistent improvements across multiple model scales on multimodal reasoning benchmarks.
We anticipate that continued scaling of process-supervised RL training will further drive fundamental advances in multimodal intelligence.

\section{Limitations}
\label{sec:limitations}

Potential failure modes include misalignment between minimal surface edits and
the true causal error spans due to paraphrasing or reasoning style shifts.
In addition, the effectiveness of Difference Feedback depends on the repair
model producing correct trajectories with relatively small edits; inaccurate or
overly large repairs may introduce noisy difference masks.
Finally, DF requires reliable correctness signals and incurs additional
inference from the repair model, which may introduce modest overhead in
large-scale training.

\section*{Ethics Statement}
This work involves no significant ethical risks.

\section*{Reproducibility Statement}
Code will be anonymously released via a public repository (GitHub) to enable
reproduction and independent verification of all results.

%
%
\bibliographystyle{splncs04}
\bibliography{main}
\end{document}